\documentclass{article}
\usepackage{newtxtext}

\PassOptionsToPackage{numbers, compress}{natbib}
\usepackage[preprint]{arxiv}

\usepackage[utf8]{inputenc} % allow utf-8 input
\usepackage[T1]{fontenc}    % use 8-bit T1 fonts
\usepackage{hyperref}       % hyperlinks
\usepackage{url}            % simple URL typesetting
\usepackage{booktabs}       % professional-quality tables
\usepackage{amsfonts}       % blackboard math symbols
\usepackage{amsmath}
\usepackage{amssymb}
\usepackage{nicefrac}       % compact symbols for 1/2, etc.
\usepackage{microtype}      % microtypography
\usepackage{xcolor}         % colors
\usepackage{enumitem}        % customize lists
\usepackage{graphicx}
\usepackage{caption}
\usepackage{subcaption}
\usepackage{adjustbox}
\usepackage{rotating}
\usepackage{multirow}
\usepackage{array}
\usepackage{listings}
\usepackage{float}
\usepackage{fancyhdr}
\newcommand{\Cols}{\mathcal{C}}
\newcommand{\Tables}{\mathcal{T}}
\newcommand{\Trans}{\mathcal{F}}
\newcommand{\Agg}{\mathcal{A}}

\newcommand{\LPred}{\widehat{L}}
\newcommand{\LGold}{L^{\star}}

\newcommand{\Mscr}{\mathcal{M}_{\textsc{scr}}}
\newcommand{\Mmod}{\mathcal{M}_{\textsc{mod}}}
\newcommand{\slice}{\texttt{SLiCE}}
\setlist[itemize]{leftmargin=2em, noitemsep, topsep=0pt}

\title{Schema Lineage Extraction at Scale: Multilingual Pipelines, Composite Evaluation, and Language-Model Benchmarks}

\author{%
  Jiaqi Yin\\
  Microsoft\\
  Redmond, WA\\
  \texttt{Jackie.Yin@microsoft.com} \\
  \And
  Yi-Wei Chen \\
  Microsoft\\
  Redmond, WA\\
  \texttt{yiweichen@microsoft.com} \\
  \AND
  Meng-Lung Lee \\
  Antra. Inc. \\
  Seattle, WA \\
  \texttt{leemenglung1012@gmail.com} \\
  \And
  Xiya Liu \\
  Microsoft\\
  Redmond, WA\\
  \texttt{Xiya.Liu@microsoft.com} \\
}
\begin{document}

\maketitle
\begin{abstract}
Enterprise data pipelines, characterized by complex transformations across multiple programming languages, often cause a semantic disconnect between original metadata and downstream data. This "semantic drift" compromises data reproducibility and governance, and impairs the utility of services like retrieval-augmented generation (RAG) and text-to-SQL systems. To address this, a novel framework is proposed for the automated extraction of fine-grained schema lineage from multilingual enterprise pipeline scripts. This method identifies four key components: source schemas, source tables, transformation logic, and aggregation operations, creating a standardized representation of data transformations.
For the rigorous evaluation of lineage quality, this paper introduces the Schema Lineage Composite Evaluation (\slice{}), a metric that assesses both structural correctness and semantic fidelity. A new benchmark is also presented, comprising 1,700 manually annotated lineages from real-world industrial scripts. Experiments were conducted with 12 language models, from 1.3B to 32B small language models (SLMs) to large language models (LLMs) like GPT-4o and GPT-4.1. The results demonstrate that the performance of schema lineage extraction scales with model size and the sophistication of prompting techniques. Specially, a 32B open-source model, using a single reasoning trace, can achieve performance comparable to the GPT series under standard prompting. This finding suggests a scalable and economical approach for deploying schema-aware agents in practical applications.
\end{abstract}

% \begin{figure}[htbp]
% \centering
% \includegraphics[width=0.8\textwidth]{figs/workflow.png}
% \caption{The workflow begins with a prompt—comprising a script, instruction, example, and schema query—which is input to an LLM or SLM. The model generates a raw text response that is then parsed into a predicted schema lineage ($\LPred$). This predicted lineage is compared against a human-annotated gold standard lineage ($\LGold$) using an evaluator. The evaluation results subsequently guide prompting engineering to iteratively refine and improve the prompt design.}
% \label{fig:framework}
% \end{figure}

\section{Introduction}
% Check and fix reference formats
% mention real data script, and human verified schema lineage, why we bring up four elements in schema lineage, RAG friendly
% 
Enterprise databases are foundational repositories powering critical business activities, including strategic decision-making, operational health monitoring, and user experience optimization. Data scientists, analysts, and engineers extensively rely on these data centers to generate actionable insights from raw data. Typically, comprehensive metadata documentation, including schema definitions and data semantics, is created alongside the initial raw datasets, providing valuable context for interpretation and utilization.

However, this initial metadata rapidly becomes outdated and ineffective as raw data undergoes extensive transformations through complex multi-stage processing pipelines. These pipelines frequently involve heterogeneous programming languages such as SQL, Python, and C\#, each employed at different processing stages for operations like data renaming, aggregation, and restructuring. Such transformations fundamentally alter the original data schemas, creating a significant semantic gap between initial metadata and derived datasets used for business intelligence, analytics dashboards, and machine learning model training.

This disconnect introduces a severe documentation gap known as "semantic drift"~\cite{Muller2016framework}, critically impeding data literacy, reproducibility, and governance within organizations. Consequently, non-technical stakeholders, analysts, and even data scientists face substantial difficulties tracing downstream metrics such as Monthly Active Users (MAU), customer churn rates, and revenue back to their precise data origins ~\cite{cui2003lineagetracing}. This reliance on a small group of technical specialists who authored the transformations, or on sparse, manually maintained documentation, severely limits the scalability of data-driven decision-making and analytics capabilities within enterprises.

While LLMs pretrained on general corpora, including metric definitions and transformation patterns, 
offer potential for automating metadata documentation~\cite{chen2021evaluatinglargelanguagemodels}, they lack access to enterprise-specific context due to strict privacy and security constraints~\cite{bommasani2022opportunitiesrisksfoundationmodels}. Even when deployed internally, third-party LLMs remain ineffective at capturing the nuanced semantics of transformed schemas without task-specific fine-tuning~\cite{lewis2021retrievalaugmentedgenerationknowledgeintensivenlp,barnett2024sevenfailurepointsengineering}.

The critical lineage information from pipeline scripts is essential to bridge semantic drift in enterprise data pipelines. We begin by formally defining schema lineage as a structured representation consisting of four essential components: source schemas, source tables, transformation logic, and aggregation operations. This compact yet expressive format captures the complete semantic path of derived schema elements from origin to output across complex, multi-language scripts.

To support benchmarking, we manually annotated 1,700 schema lineages across 50 real-world enterprise pipeline scripts written in SQL, Python, and C\#. These scripts span diverse business domains and complexity levels, offering a high-fidelity benchmark for schema lineage extraction. We introduce \textbf{\slice{}} (Schema Lineage Composite Evaluation) to quantify the extraction accuracy. \slice{} is a novel metric that combines structural validity and semantic correctness into a unified score between 0 and 1, while exposing component-level diagnostics across format, source schemas, source tables, transformation logic, and aggregation.

We conduct extensive experiments across 12 language models, including SLMs ranging from 1.3B to 32B parameters and LLMs such as GPT-4o and GPT-4.1. Our evaluation spans three prompting strategies, base, few-shot, and chain-of-thought. It reveals key trends on how model scale, prompt design, and script complexity affect schema lineage extraction quality. Notably, we demonstrate that a 32B open-source model, when guided by a single reasoning trace, achieves performance comparable to GPT-series model, offering a cost-effective path for industrial deployment.

In summary, we make four key contributions: (1) a formal definition of schema lineage tailored to multi-language enterprise pipelines, capturing source-to-output semantics across transformation logic and aggregation; (2) a high-quality benchmark of 1,700 manually annotated schema lineages from 50 real-world scripts; (3) the \slice{} metric, a comprehensive evaluation framework that enables fine-grained assessment of extraction quality; and (4) extensive experiments across 12 language models, demonstrating how model scale, prompting strategy, and script complexity influence extraction performance.

% Through automated schema lineage extraction, which empowers RAG systems, text-to-SQL generation frameworks~\cite{yu2019spiderlargescalehumanlabeleddataset}, and broader enterprise data analytics by ensuring accurate, dynamic, and contextually rich schema documentation that seamlessly evolves alongside the underlying data transformation processes.

\section{Dataset and Schema Lineage Definition}

\subsection{Enterprise Data Pipeline Collection}

Industries routinely employ sophisticated, multi-stage, and multi-language transformation pipelines to support diverse analytical workflows. These pipelines typically begin with large-scale preprocessing using frameworks like PySpark and Scope~\cite{zhou2012scope} and transition to downstream metric computation in SQL or Python, reflecting the heterogeneity of real-world data engineering environments.

To capture these complexities, we curated a comprehensive dataset comprising 50 representative enterprise data pipeline scripts. These scripts span multiple programming languages, including SQL, C\#, and Python (including PySpark). Each script, actively deployed within Microsoft, serves distinct analytical purposes, ranging from business metrics computation to marketing analytics, product insights, and user experience optimization. 

We categorized scripts into three difficulty levels (easy, medium, hard) based on quantitative criteria detailed in Appendix~\ref{sec:script-complexity}.  Our dataset includes 19 easy scripts (averaging 26 schemas and 921 tokens per script), 22 medium scripts (averaging 28 schemas and 1,806 tokens per script), and 9 hard scripts (averaging 67 schemas and 4,687 tokens per script), collectively amounting to 1,700 schema annotations (Table \ref{tab:dataset_stats}).

\begin{table}[ht]
\centering
\caption{Overview of enterprise data pipeline scripts categorized by complexity level, detailing token count and schema statistics}
\label{tab:dataset_stats}
\small
\begin{tabular}{@{}lcccccccc@{}}
\toprule
\textbf{Difficulty} & \textbf{Scripts} & \multicolumn{3}{c}{\textbf{Token Count}} & \multicolumn{4}{c}{\textbf{Schema Count}} \\
\cmidrule(lr){3-5} \cmidrule(lr){6-9}
& & \textbf{Avg.} & \textbf{Min} & \textbf{Max} & \textbf{Total} & \textbf{Avg.} & \textbf{Min} & \textbf{Max} \\
\midrule
All & 50 & 1,988.52 & 139 & 17,447 & 1,700 & 34.00 & 5 & 391 \\
\midrule
Easy & 19 & 921.26 & 139 & 2,153 & 488 & 25.68 & 5 & 118 \\
Medium & 22 & 1,806.23 & 274 & 6,882 & 610 & 27.73 & 6 & 109 \\
Hard & 9 & 4,687.22 & 751 & 17,447 & 602 & 66.89 & 10 & 391 \\
\bottomrule
\end{tabular}
\end{table}

While the full dataset is based on real, production-level scripts, we simulate a hard example in Appendix \ref{section: script} to illustrate their structural and logical characteristics. These examples preserve the multi-stage, multi-language complexity of the original pipelines while changing sensitive business logic.

\subsection{Schema Lineage Definition and Annotation}
\label{sec:schema-lineage}

\begin{figure}[htbp]
\centering
\includegraphics[width=1.0\textwidth]{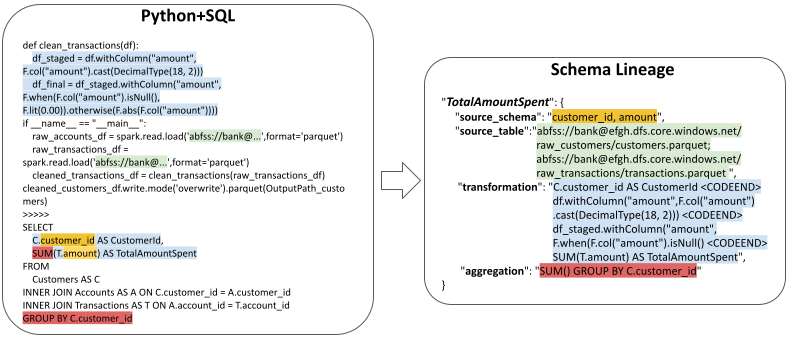}
\caption{A visual illustration of schema lineage definition and annotation, based on the formal structure introduced in Section \ref{sec:schema-lineage}. The example demonstrates how raw data pipeline scripts combined with Python and SQL code is analyzed to extract the four core components (source schemas, source tables, transformation logic, and aggregation operations) of schema lineage for \texttt{TotalAmountSpent}. The resulting structured lineage represents a human-labeled gold annotation used for model evaluation and training.}
\label{fig:schema-lineage}
\end{figure}

We formally define schema lineage as a structured representation capturing the semantics of data transformations within enterprise data pipelines. 
A typical pipeline script reads from one or more source tables and produces one output table as a result of transformation logic. The output table consists of multiple schemas, corresponding to the columns or fields.
For each schema in an output table, 
we extract a distinct schema lineage that traces its derivation from the original data sources. 
We conceptualize schema lineage as a structured mapping comprising four essential components:
\begin{itemize}
  \item \textbf{Source Schemas}: The original schema elements from which lineage originates, indicating the foundational data fields contributing to the resultant schema. Multiple source schemas, if applicable, are comma-separated.
  \item \textbf{Source Tables}: The initial data tables containing source schemas, acting as primary data origins.
  \item \textbf{Transformation}: The explicit code snippet or operational logic applied to transform source schemas into the resultant schema. A sequence of transformations is delimited using the \texttt{<CODEEND>} separator.
  \item \textbf{Aggregation}: Aggregation operations applied throughout transformation, such as \texttt{GROUP BY}, \texttt{SUM}, \texttt{COUNT}, \texttt{MAX}, or \texttt{MIN}, alongside their grouping keys. A sequence of aggregations is similarly separated using the \texttt{<CODEEND>} delimiter.
\end{itemize}
Those components are essential because schema lineage serves as the connective tissue between raw data and downstream outputs.  Without understanding how each schema element was derived, it becomes impossible to reconstruct the full context of a dataset, explain business metrics, or enable AI agents to operate reliably. For instance, tracing the lineage of a metric like \texttt{TotalAmountSpent} showed in Figure \ref{fig:schema-lineage} requires more than matching column names. It demands precise reconstruction of how those values were computed, transformed, and aggregated from their original tables.

Our dataset includes schema lineages manually annotated by human experts, following strict consistency guidelines to ensure reliable experimental evaluation. Through meticulous annotation, we have produced 1,700 high-quality schema lineages covering diverse complexity levels and transformation patterns, creating a robust gold standard for evaluating language models.
Moreover, detailed reasoning traces were generated for each script to support varied prompting strategies during evaluation; seen in Section~\ref{sec:prompting}. These traces strengthen our assessment framework, ensuring comprehensive and rigorous evaluations of automated schema lineage extraction.

\section{Schema Lineage Composite Evaluation (\slice{})}
%The automatic extraction of schema lineage from enterprise data pipelines presents unique challenges that distinguish it from conventional code analysis tasks. The traditional evaluation metrics access correctness through execution outcomes (\texttt{pass@k})~\cite{chen2021evaluatinglargelanguagemodels} or syntactic  similarity ~\cite{ren2020codebleumethodautomaticevaluation}. 
Schema lineage requires  accurately identifying the original source columns, tracing transformation logic, and capturing aggregation operations, even when they are distributed across multiple abstraction layers or languages. To meet these requirements, we propose a novel evaluation metrics called, \textbf{\slice{}}, specifically designed for \textbf{S}chema \textbf{Li}neage \textbf{C}omposite \textbf{E}valuation.  Our approach recognizes that successful lineage extraction must satisfy multiple criteria simultaneously: structural correctness, semantic accuracy, and practical utility for enterprise applications. 

\subsection{Problem Statement}

Given an enterprise data pipeline script and a target schema from the output table, our objective is to extract the corresponding schema lineage that traces the data transformation process from source to target. 
Let $\mathcal{S}$ represent a data pipeline script of multiple programming languages (SQL, C\#, Python), and let $\sigma$ denote a target schema in the output table generated by $\mathcal{S}$. Our goal is to map the script-schema pair to a structured schema lineage $L$.
Each schema lineage $L$ is defined as a structured dictionary with four required keys: \texttt{source\_schema, source\_table, transformation, aggregation}.
% \begin{verbatim}
% {
%     "source_schema": "...",
%     "source_table": "...", 
%     "transformation": "...",
%     "aggregation": "..."
% }
% \end{verbatim}
This key-value format is essential for both evaluation and downstream parsing.

To streamline our mathematical formulation, we denote the value corresponding to each key using the following symbols: $\Cols$ for \texttt{source\_schema} representing the set of source columns,
$\Tables$ for \texttt{source\_table} denoting the set of tables,
$\Trans$ for \texttt{transformation}, the transformation logic, and $\Agg$ for \texttt{aggregation}, the final aggregation expression.
We thus represent the schema lineage as a structured mapping:
\begin{align*}
L = \left\{ \begin{aligned}
    &\texttt{source\_schema}: \Cols,\\
    &\texttt{source\_table}: \Tables,\\
    &\texttt{transformation}: \Trans,\\
    &\texttt{aggregation}: \Agg
\end{aligned} \right\}.
\end{align*}
During evaluation, we consider a predicted lineage $\LPred$ generated by a language model and a gold standard lineage $\LGold$ annotated by experts.
% \[
% \begin{array}{c@{\qquad}c}
% \LPred = \left\{ \begin{aligned}
%     &\texttt{source\_schema}: \widehat{\Cols},\\
%     &\texttt{source\_table}: \widehat{\Tables},\\
%     &\texttt{transformation}: \widehat{\Trans},\\
%     &\texttt{aggregation}: \widehat{\Agg}
% \end{aligned} \right\}
% &
% \LGold = \left\{ \begin{aligned}
%     &\texttt{source\_schema}: \Cols^\star,\\
%     &\texttt{source\_table}: \Tables^\star,\\
%     &\texttt{transformation}: \Trans^\star,\\
%     &\texttt{aggregation}: \Agg^\star
% \end{aligned} \right\}
% \end{array}
% \]
This dictionary-based representation ensures alignment with both the model's output structure and the evaluation interface.

\subsection{\slice{} Definition}\label{sec:metrics} 
The \slice{} integrates structural validity~\cite{deepseekai2025deepseekr1incentivizingreasoningcapability} and semantic correctness~\cite{ren2020codebleumethodautomaticevaluation}
into a single value $\slice(\LPred,\LGold)\!\in[0,1]$, while still exposing
component-level diagnostics (format, source, tables, transformation, aggregation).

% Our evaluation decomposes lineage quality into four distinct components, allowing fine-grained analysis of model behavior across different aspects of lineage generation.

\paragraph{Format Correctness.} 
Schema lineage extraction requires strict adherence to both the dictionary structure and the output scaffolding imposed by the prompting strategy. Depending on whether the model is prompted with or without intermediate reasoning, the response must conform to one of the following formats:
\begin{itemize}
    \item \textbf{With reasoning trace:} the response must contain both reasoning and answer blocks:
    \begin{verbatim}
<think> ... reasoning trace ... </think>
<answer> ... schema lineage dictionary ... </answer>
    \end{verbatim}
    \item \textbf{Without reasoning trace:} the response must contain only the answer block:
    \begin{verbatim}
<answer> ... schema lineage dictionary ... </answer>
    \end{verbatim}
\end{itemize}
In both cases, the lineage content inside the \texttt{<answer>} \texttt{</answer>} tag must follow a strict key-value dictionary format, containing exactly four keys: \texttt{source\_schema}, \texttt{source\_table}, \texttt{transformation}, and \texttt{aggregation}.

The format correctness score enforces all these format constraints jointly:
\begin{equation}
\mathcal{M}_{\textsc{fmt}}(\LPred) = 
\begin{cases}
1 & \text{if } \LPred \text{ satisfies all \texttt{<tag>} structure and dictionary key requirements} \\
0 & \text{otherwise}
\end{cases}.
\label{eq:format-correctness}
\end{equation}
Any deviation, such as malformed tags, incorrect key names, or missing fields , results in immediate failure. This reflects the importance of strict structural adherence in automated parsing systems.

\paragraph{Source Schema Evaluation.}  
Source schemas represent the foundational elements of lineage extraction, specifying which original columns contribute to the target schema. We compute a binary match based on exact set equality:
\begin{equation}
\mathcal{M}_{\textsc{src}}(\LPred, \LGold) = 
\begin{cases}
1 & \text{if } \widehat{\Cols} = \Cols^\star \\
0 & \text{otherwise}
\end{cases}.
\label{eq:source-schema}
\end{equation}
This metric enforces strict case-sensitive, order-insensitive matching of column names.

\paragraph{Source Table Evaluation.}  
There are variations in source table naming conventions and hierarchies (e.g., database.schema.table vs. table), a simple exact-match metric is insufficient. $\mathcal{M}_{\textsc{tbl}}$ is proposed to combine a strict exact-match $F1$ score with a more flexible fuzzy similarity score $F_u$:
\begin{equation}
\mathcal{M}_{\textsc{tbl}}(\LPred, \LGold) = 
w_1^{\textsc{tbl}} \cdot F1(\widehat{\Tables}, \Tables^\star) + 
w_2^{\textsc{tbl}} \cdot F_u(\widehat{\Tables}, \Tables^\star),
\label{eq:source-tables}
\end{equation}
where the weights $w_1^{\textsc{tbl}} + w_2^{\textsc{tbl}} = 1$. $F_u$ is defined to provides partial credit for predictions that are textually similar but not identical to the ground truth:
\begin{equation}
F_u(\widehat{\Tables}, \Tables^\star) = \frac{1}{2} \left[
\frac{1}{|\widehat{\Tables}|} \sum_{t_i \in \widehat{\Tables}} \max_{t_j \in \Tables^\star} \text{FuzzyMatch}(t_i, t_j)
+
\frac{1}{|\Tables^\star|} \sum_{t_j \in \Tables^\star} \max_{t_i \in \widehat{\Tables}} \text{FuzzyMatch}(t_i, t_j)
\right].
\label{eq:fuzzy-match}
\end{equation}
 The $\text{FuzzyMatch}(t_i, t_j)$ function computes a normalized similarity ratio of table name based on the Levenshtein distance~\cite{levenshtein1966binary}. The first term in Equation \ref{eq:fuzzy-match}, Fuzzy precision, measures how well each predicted table matches the best candidate in the ground-truth set, while the second term, Fuzzy recall, measures how well each ground-truth table is represented by its best match in the predicted set. The \texttt{max} operator ensures a table is only scored against its most similar counterpart. This hybrid approach provides a more nuanced evaluation that rewards exactness while accommodating common naming variations.

\paragraph{Transformation and Aggregation Evaluation.}  
Transformation ($\Trans$) and aggregation ($\Agg$) fields contain code snippets in various programming languages. These components are challenging to evaluate due to the possibility of logical equivalence despite syntactic variation. 
We define a novel Multi-AST similarity in Eq.~\ref{eq:multi-ast} that supports multilingual code comparison. 
First, we compute language-aware AST similarity:
\begin{equation}
\text{AST}_{\text{multi}}(\widehat{x}, x^\star) = \sum_{l \in \mathcal{L}} w_l \cdot \text{AST}_l(\widehat{x}, x^\star)
\label{eq:multi-ast},
\end{equation}
where $x \in \{\Trans, \Agg\}$, $\mathcal{L}$ is the set of candidate languages, and $w_l$ is the confidence that $x$ belongs to language $l$ (with $\sum_l w_l = 1$). The weight $w_l$ is computed as the normalized proportion of language-specific keywords observed in $x$, serving as a proxy for language attribution.

Motivated by CodeBLEU~\cite{ren2020codebleumethodautomaticevaluation}, we define the component metrics for transformation and aggregation, repectively, as a weighted average of standard BLEU~\cite{papineni-etal-2002-bleu}, weighted BLEU (as introduced in CodeBLEU), and a modified AST-based similarity:
\begin{align}
\mathcal{M}_{\textsc{trf}}(\LPred, \LGold) &= 
w_1^{\textsc{trf}} \cdot \text{BLEU}(\widehat{\Trans}, \Trans^\star)
+ w_2^{\textsc{trf}} \cdot \text{BLEU}_{\text{weight}}(\widehat{\Trans}, \Trans^\star)
+ w_3^{\textsc{trf}} \cdot \text{AST}_{\text{multi}}(\widehat{\Trans}, \Trans^\star),
\label{eq:transformation}
\\
\mathcal{M}_{\textsc{agg}}(\LPred, \LGold) &= 
w_1^{\textsc{agg}} \cdot \text{BLEU}(\widehat{\Agg}, \Agg^\star)
+ w_2^{\textsc{agg}} \cdot \text{BLEU}_{\text{weight}}(\widehat{\Agg}, \Agg^\star)
+ w_3^{\textsc{agg}} \cdot \text{AST}_{\text{multi}}(\widehat{\Agg}, \Agg^\star).
\label{eq:aggregation}
\end{align}
Each set of weights satisfies $\sum_{i=1}^3 w_i^{\textsc{trf}} = 1$ and $\sum_{i=1}^3 w_i^{\textsc{agg}} = 1$.
While CodeBLEU includes AST similarity for single-language code, our approach extends this term to support multi-language settings by computing a language-aware aggregation over candidate AST parsers. We exclude the data-flow matching term from CodeBLEU, as schema lineage transformations often consist of partial and non-executable code snippets.

\paragraph{Composite Performance Score.}  
% We integrate all metrics through a hierarchical strategy that enforces structural validity (format and source schema) as a prerequisite for semantic evaluation. 
The final evaluation metric is defined as:
\begin{equation}
\slice(\LPred, \LGold) = \mathcal{M}_{\textsc{fmt}}(\LPred) \cdot \mathcal{M}_{\textsc{src}}(\LPred, \LGold) \cdot
\left[
\omega_{\textsc{tbl}} \cdot \mathcal{M}_{\textsc{tbl}} +
\omega_{\textsc{trf}} \cdot \mathcal{M}_{\textsc{trf}} +
\omega_{\textsc{agg}} \cdot \mathcal{M}_{\textsc{agg}}
\right],
\label{eq:slice-score}
\end{equation}
where the weights are predefined and satisfy $\omega_{\textsc{tbl}} + \omega_{\textsc{trf}} + \omega_{\textsc{agg}} = 1$. 
Other weights $(w_1^{\textsc{tbl}}, w_2^{\textsc{tbl}})$ in $M_{\textsc{tbl}}$, $(w_1^{\textsc{trf}}, w_2^{\textsc{trf}}, w_3^{\textsc{trf}})$ in $M_{\textsc{trf}}$, $(w_1^{\textsc{agg}}, w_2^{\textsc{agg}}, w_3^{\textsc{agg}})$ in $M_{\textsc{agg}}$ are also predefined.
Note that $\mathcal{M}_{\textsc{fmt}}$ and $\mathcal{M}_{\textsc{src}}$ are binary values. Eq.~\ref{eq:slice-score} ensures that violations in basic structural constraints (e.g., incorrect format or source columns) nullify downstream correctness, reflecting how such errors propagate through real-world systems.

% This evaluation framework supports both holistic and component-wise assessment, enabling targeted diagnosis of model performance in schema lineage extraction.
The proposed metric, \slice{}, offers a principled foundation for systematic performance analysis and model diagnostics. The fine-grained, component-wise scoring enables detailed benchmarking of language model capabilities across distinct aspects of schema lineage extraction, as demonstrated in our experiments. Importantly, the structured formulation of the \slice{} metric is not only useful for evaluation but also well-suited to serve as a reward signal in future supervised fine-tuning or reinforcement learning frameworks~\cite{shojaee2023executionbasedcodegenerationusing, deepseekai2025deepseekr1incentivizingreasoningcapability}. 
% While our current work focuses on evaluation, this extensibility highlights the broader potential of \slice{} as both a rigorous assessment tool and a building block for advancing model alignment in enterprise data pipeline understanding.

\subsection{Prompting Categories}\label{sec:prompting}
To systematically investigate the level of contextual richness on performance of schema lineage extraction, we design three hierarchical prompting categories. Their detailed prompt examples can be found in Section \ref{app-sec:prompts}.
\begin{itemize} 
\item \textbf{Base Prompting}: This strategy provides only the essential pipeline script along with explicit extraction instructions specifying target output formats and component definitions. It serves as a baseline by representing the minimal necessary context required for schema lineage extraction.

\item\textbf{Few-Shot Prompting}: This strategy enhances the base prompting approach by integrating concrete input-output example pairs directly into the prompt, providing tangible references that guide the language model's understanding of expected outputs. We scale the quantity of these examples according to pipeline complexity, providing one example for easy pipelines, up to two for medium-complexity pipelines, and up to three for hard pipelines.

\item \textbf{Chain-of-Thought (CoT)}: Building upon few-shot prompting, this advanced strategy incorporates detailed human-generated reasoning traces that illustrate step-by-step derivations of schema lineage from pipeline code. The inclusion of explicit reasoning processes aims to guide the language model through logical inference steps.
% thereby enhancing its semantic interpretation and extraction accuracy.
\end{itemize}

% To improve inference efficiency during evaluation, 
Note that we apply PagedAttention~\cite{kwon2023efficientmemorymanagementlarge}, a key-value caching mechanism, for open source small language models. It virtualizes the key-value cache memory to prevent fragmentation and optimize reuse. Since the constructed prompts, including scripts, extraction instructions, and provided examples, remain invariant for different schema queries within the same pipeline script, we compute and cache the model’s key-value pairs once per pipeline. This optimization significantly reduces redundant computational efforts and accelerates the schema lineage extraction process.

\section{Related Work}
Early approaches to schema lineage extraction primarily relied on conventional code analysis techniques, including abstract syntax tree (AST) parsing~\cite{sqlparse, sqlfluff}, metadata mapping~\cite{mslearn_purview_classic_lineage}, and runtime analysis~\cite{foundational_data_lineage}. These methods can achieve reliable results in single-language, static environments, but they often struggle to scale when faced with the complexity of multi-stage, multi-language data pipeline. The need to continuously adapt to evolving codebases and heterogeneous scripts further limits their applicability.

More recently, advances in large language models (LLMs) have opened new directions for automated lineage extraction~\cite{li2025llm_data_lineage_parsing, kuznetsov2025_llim_cyberhaven}. By reframing lineage extraction as a code understanding task, LLM-based methods have demonstrated strong performance. Chain-of-Thought (CoT) prompting, combined with a handful of examples~\cite{li2025llm_data_lineage_parsing}, enables LLMs to generate high-quality schema lineages, 
even without task-specific fine-tuning. However, existing CoT approaches typically generate table- and operation-level lineages separately, missing opportunities for comprehensive extraction.
Fine-tuned solutions, such as LLiM~\cite{kuznetsov2025_llim_cyberhaven}, further leverage enterprise signals by converting customer lineage data into positive and negative event labels. These models effective at capturing anomalous lineage patterns, but they may lack the generalization to answer more fundamental schema dependency questions.
Our work advanced the no-fine-tuning paradigm, where LLMs simultaneously generate both table- and operation-level lineages in a single query. Our dataset is curated for data understanding, rather than downstream applications such as anomaly detection. This setup allows for a more precise assessment of schema lineage extraction in realistic pipeline scenarios.

Open lineage datasets are extremely rare, as lineages often encode sensitive business logic and confidential data relationships. Benchmarks like TPC-H~\cite{apache_doris_tpch_benchmark} serve as templates for synthesizing data processing scripts with LLM~\cite{li2025llm_data_lineage_parsing}. These synthesized scripts typically use a single language, such as SQL or Python, and lack the complexity of hundred-line and multi-language implementations. 
In contrast, data lineage graphs constructed from enterprise applications~\cite{chen2024_data_lineage_graphs} more accurately reflect real-world data dependency. These graphs represent tables, SQL code snippets, and database columns as nodes, with edges denoting SQL transformation, making them effective for evaluating data provenance techniques. However, they are not tailored for LLM-based lineage extraction benchmarks.
Our dataset is composed of selected real-word scripts spanning multiple languages. It inherits the structural benefits of lineage graphs~\cite{chen2024_data_lineage_graphs} and is explicitly designed to support retrieval augmentation generation (RAG)~\cite{Liu_LlamaIndex_2022} and  Text-to-SQL applications~\cite{yu2019spiderlargescalehumanlabeleddataset}.

Existing evaluation metrics for code generation typically fall into two categories: execution-based outcomes, notably \texttt{pass@k}~\cite{chen2021evaluatinglargelanguagemodels}, and semantic matching metrics, such as CodeBLEU~\cite{ren2020codebleumethodautomaticevaluation}. 
% Given that lineage extraction can be framed as a code generation task, these metrics offer natural baselines for evaluation. 
While lineage extraction naturally aligns with code generation paradigms, conventional metrics present fundamental limitations in this domain.
The \texttt{pass@k} metric estimates the probability that, out of n generated code samples, at least one of k randomly selected outputs passes all test cases. 
The lineage extraction tasks that inherently produce partial transformation and aggregation logic violate the the requirement of complete executable programs for the \texttt{pass@k}.
% However, because lineage extraction typically yields partial transformation or aggregation logic rather than fully executable programs, adopting \texttt{pass@k} is not feasible. 
CodeBLEU is a composite score incorporating n-gram BLEU scores~\cite{papineni-etal-2002-bleu}, weighted n‑gram match, AST similarity, and data-flow semantic matching.
% the transformation logic in generated lineages often comprises incomplete code fragments and may involve multiple programming languages. They makes data-flow analysis and monolingual AST matching inapplicable. 
Nevertheless, 
its applicability remains constrained by two critical factors: the heterogeneous nature of transformation logic with multiple programming languages, and the prevalence of syntactically incomplete code fragments that preclude traditional AST and data-flow analyses
Furthermose, the output format of LLM-generated lineage is a critical consideration for fine-tuning~\cite{deepseekai2025deepseekr1incentivizingreasoningcapability}. Our proposed \slice{} metric preserves CodeBLEU's theoretical foundations while providing native support for partial, multilingual code evaluation and incorporating contemporary LLM fine-tuning considerations~\cite{deepseekai2025deepseekr1incentivizingreasoningcapability}.

\section{Experiments}\label{sec:experiments}
Our experimental evaluation is designed to investigate several key aspects of schema lineage extraction. We compare the performance of state-of-the-art LLMs with specialized SLMs to understand their relative capabilities on this task across data pipelines of varying difficulty.
% We analyze the relationship between model size and extraction performance  
Methodologically, we assess the impact of different prompting strategies on extraction accuracy and validate the effectiveness of our proposed lineage metrics.

%State-of-the-art LLMs such as GPT-4o and GPT-4.1 \citep{openai2025gpt41} set a performance ceiling, however their direct application to schema lineage tasks poses significant cost and efficiency challenges. These APIs require complete pipeline scripts and extraction instructions in every request, resulting in high token consumption proportional to the number of schemas and scripts. 

%Thus, we also evaluate specialized SLMs aligned with coding tasks to assess their viability in practical deployments.

\subsection{Model Selection and Experimental Setup}
Initially, we evaluated a comprehensive set of language models, encompassing two LLMs (GPT-4.1\cite{openai2025gpt41} and GPT-4o~\cite{openai_gpt4o_2024}), alongside 16 distinct  SLMs. 
The SLM cohort included Qwen2.5-Coder variants (1.5B, 3B, 7B, 14B, 32B)~\cite{hui2024qwen25codertechnicalreport}, Mistral-7B~\cite{jiang2023mistral7b}, Codestral-22B~\cite{mistral_codestral_2024}, CodeLlama variants (7B, 13B, 34B)~\cite{rozière2024codellamaopenfoundation}, DeepSeek-Coder variants (1.3B, 6.7B, 16B))~\cite{deepseekai2024deepseekcoderv2breakingbarrierclosedsource}, and Phi-4 configurations (mini~\cite{xu2025phi4minireasoningexploringlimitssmall}, 14B~\cite{abdin2024phi4technicalreport}, reasoning-14B~\cite{abdin2025phi4reasoningtechnicalreport}). 
The majority of these models underwent pretraining on code corpora or subsequent alignment for coding tasks, establishing their reputation for robust coding capabilities \citep{jiang2024surveylargelanguagemodels}. 
% A subset, including 
Mistral-7B~\cite{jiang2023mistral7b} and Phi-4~\cite{abdin2024phi4technicalreport} series represents general-purpose architectures to assess the performance characteristics of domain-agnostic SLMs in coding contexts.
After preliminary assessments, we excluded six models due to either excessive inference time or consistently poor performance, resulting in the final selection: Qwen2.5-Coder (1.5B, 3B, 7B, 14B, 32B), Mistral-7B, Codestral-22B, DeepSeek-Coder (1.3B, 6.7B, and Phi-4 (14B).

We extract schema lineage across 50 data pipeline scripts using three categories of prompting strategies detailed in Section \ref{sec:prompting}. Data experts crafted human reasoning traces to support the CoT prompting strategy: one reasoning trace per easy script, two per medium script, and three per hard script. Consequently, we implemented seven distinct prompting strategies: base, few-shot with one example (one-shot), few-shot with two examples (two-shot), few-shot with three examples (three-shot), CoT with one reasoning trace (CoT-1), CoT with two reasoning traces (CoT-2), and CoT with three reasoning traces (CoT-3). This comprehensive design allows us to investigate how different prompting strategies influence extraction accuracy across varying script complexities.
We parse these outputs and evaluate the predicted schema lineage ($\LPred$) against expert-annotated ground truth ($\LGold$) using \slice{} scores. The weights of \slice{} are assigned for the all experiments,
$w_1^{\textsc{tbl}}=0.7, w_2^{\textsc{tbl}}=0.3; w_1^{\textsc{trf}}=w_1^{\textsc{agg}}=0.5, w_2^{\textsc{trf}}=w_2^{\textsc{agg}}=0.3, 
w_3^{\textsc{trf}}=w_3^{\textsc{agg}}=0.2; 
\omega_{\textsc{tbl}}=0.4, \omega_{\textsc{trf}}=0.4, \omega_{\textsc{agg}}=0.2$.

\paragraph{Evaluation Protocol.}
For each language model $\Theta$, schema lineage is extracted across all target schemas within the 50 scripts using the prompting strategies defined in Section~\ref{sec:prompting}. Model predictions are scored against expert annotations using the \slice{} metric defined in Section~\ref{sec:metrics}. For each script $s_i$ containing schemas $\sigma_{ik}$, we compute a \textit{script-level} score by averaging schema-level scores:
\begin{equation}
\Mscr(s_i,\Theta)=\frac{1}{K_i}\sum_{k=1}^{K_i}\slice \bigl(\LPred_{ik},\,\LGold_{ik}\bigr),
\label{eq:script-score}
\end{equation}
where $\LPred_{ik}$ and $\LGold_{ik}$ represent predicted and gold lineage, respectively. To derive a \textit{corpus-level} evaluation, we average across all scripts:
\begin{equation}
\Mmod(\Theta)=\frac{1}{I}\sum_{i=1}^{I}\Mscr(s_i,\Theta) = \frac{1}{I}\sum_{i=1}^{I}\frac{1}{K_i}\sum_{k=1}^{K_i}\slice \bigl(\LPred_{ik},\,\LGold_{ik}\bigr).
\label{eq:overall-score-model}
\end{equation}

% \noindent\textbf{Stability Analysis.}
% The entire experimental process is replicated six times with different random seeds. We report the mean and standard deviation of $\Mmod(\Theta)$, providing insights into metric stability and variability of model performance.

\paragraph{Experimental Design Summary.}
Our evaluation involves 12 language models (two LLMs and 10 SLMs) across 50 data pipeline scripts. We employ three core prompting categories (base, few-shot, and CoT) resulting in 7 strategies adjusted by script complexity.
% : easy pipelines receive one example, medium pipelines receive one to two examples, and hard pipelines receive up to three examples. 
The experimental framework includes six randomized trials resulting in over 50,000 individual extraction tasks across all the conditions. 
We report the mean and standard deviation of $\Mmod(\Theta)$, providing insights into metric stability and variability of model performance.
% Source_table_weights={'f1': 0.7, 'fuzzy': 0.3},
%                  transformation_weights={'bleu': 0.5, 'weighted_bleu': 0.3, 'ast': 0.2},
%                  aggregation_weights={'bleu': 0.5, 'weighted_bleu': 0.3, 'ast': 0.2},
% LineageComponents.SOURCE_TABLE: 0.4,
%                 LineageComponents.TRANSFORMATION: 0.4,
%                 LineageComponents.AGGREGATION: 0.2
\begin{table}[ht]
\centering
\caption{Benchmark results of 12 language models evaluated on schema lineage extraction from 50 data pipeline scripts using three prompting strategies: base (zero-shot), one-shot, and chain-of-thought with a single reasoning trace (CoT-1). Mean corpus-level \slice{} scores and standard deviations are reported across six random seeds, ordered by model size.}
\label{tab:model-performance}
\begin{tabular}{@{}lcccc@{}}
\toprule
\textbf{Model} & \textbf{Size} & \textbf{Base} &\textbf{One-Shot} & \textbf{CoT-1} \\
\midrule
\multicolumn{4}{@{}l}{\textit{LLMs}} \\[2pt]
GPT-4.1~\cite{openai2025gpt41} & - & 0.418 $\pm$ 0.005 & 0.673 $\pm$ 0.008 & 0.767 $\pm$ 0.007 \\
GPT-4o~\cite{openai_gpt4o_2024} & - & 0.284 $\pm$ 0.003 & 0.654 $\pm$ 0.007 & 0.759 $\pm$ 0.008 \\
\midrule
\multicolumn{4}{@{}l}{\textit{SLMs}} \\[2pt]
 DeepSeek-Coder~\cite{deepseekai2024deepseekcoderv2breakingbarrierclosedsource}& 1.3B & 0.000 $\pm$ 0.000 & 0.054 $\pm$ 0.015 & 0.038 $\pm$ 0.017 \\
 Qwen2.5-Coder~\cite{hui2024qwen25codertechnicalreport} &1.5B & 0.014 $\pm$ 0.002 & 0.309 $\pm$ 0.006 & 0.304 $\pm$ 0.017 \\
 Qwen2.5-Coder~\cite{hui2024qwen25codertechnicalreport} &3B & 0.100 $\pm$ 0.004 & 0.391 $\pm$ 0.015 & 0.445 $\pm$ 0.010 \\
 DeepSeek-Coder~\cite{deepseekai2024deepseekcoderv2breakingbarrierclosedsource} & 6.7B & 0.003 $\pm$ 0.003 & 0.084 $\pm$ 0.018 & 0.509 $\pm$ 0.007 \\
 Mistral~\cite{jiang2023mistral7b} &7B & 0.026 $\pm$ 0.003 & 0.331 $\pm$ 0.005 & 0.227 $\pm$ 0.009 \\
 Qwen2.5-Coder~\cite{hui2024qwen25codertechnicalreport} &7B & 0.167 $\pm$ 0.005 & 0.487 $\pm$ 0.018 & 0.556 $\pm$ 0.009 \\
 Phi-4~\cite{abdin2024phi4technicalreport} &14B & 0.016 $\pm$ 0.003 & 0.511 $\pm$ 0.005 & 0.648 $\pm$ 0.005 \\
 Qwen2.5-Coder~\cite{hui2024qwen25codertechnicalreport} &14B & 0.286 $\pm$ 0.004 & 0.547 $\pm$ 0.005 & 0.646 $\pm$ 0.007 \\
 Codestral~\cite{mistral_codestral_2024} &22B & 0.126 $\pm$ 0.004 & 0.511 $\pm$ 0.005 & 0.662 $\pm$ 0.008 \\
 Qwen2.5-Coder~\cite{hui2024qwen25codertechnicalreport} & 32B & 0.355 $\pm$ 0.004 & 0.623 $\pm$ 0.004 & 0.734 $\pm$ 0.007 \\
\bottomrule
\end{tabular}
\end{table}

\subsection{Results}
Table \ref{tab:model-performance} presents corpus-level performance across 12 language models, evaluated using three prompting strategies: base (zero-shot), one-shot, and chain-of-thought with one reasoning trace (CoT-1). The consistently low standard deviation observed across random seeds underscores the robustness and reliability of our evaluation metrics.

Several key patterns emerge from our results. First, base prompting consistently yields the lowest performance across all models. Introducing a single output example (one-shot) substantially improves extraction accuracy. For instance, GPT-4.1 improves its \slice{} score by 61\%, and the SLM Qwen2.5-Coder-32B sees a 75\% increase. Interestingly, general-purpose language models such as GPT-4o, Mistral, and Phi-4 exhibit even greater results with one-shot prompting, achieving improvements exceeding 100\%. 
Adding a reasoning trace (CoT-1) further enhances performance by over another 10\% for models with size $\geq$ 3B, demonstrating the effectiveness of CoT reasoning in guiding schema lineage extraction.

Secondly, there is positive correlation between model size and extraction performance. Within the same model families, holding the prompting strategy unchanged, larger models consistently outperform smaller models, highlighting model size as a significant factor; as seen in Figure~\ref{fig:model_performance_vs_param_size}.  Under CoT-1 prompting, Qwen2.5-Coder-32B achieves the highest \slice{} score of 0.734, more than doubling the performance of its smallest variant (1.5B), which scores 0.304.

We observe that CoT prompting yields diminishing returns for models with fewer than 3B parameters. For instance, DeepSeek-Coder-1.3B and Qwen2.5-Coder-1.5B exhibit decreased lineage extraction performance when moving from one-shot to CoT-1 prompting. Two potential explanations account for this trend. First, chain-of-thought reasoning is widely considered an emergent capability that typically arises in larger models, aligning with prior findings by \citet{wei2023chainofthoughtpromptingelicitsreasoning}. Second, the longer prompt lengths inherent to CoT may overwhelm small models with limited context windows. This degradation is consistent with observations by \citet{liu-etal-2024-lost}.
%For models with parameters $\geq$ 3B, improvements from one-shot to CoT-1 positively correlate with model scale.
%General-purpose models such as Mistral-7B and Phi-4 exhibit distinct behaviors. While CoT-1 prompting does not enhance Mistral-7B's performance, Phi-4 significantly benefits from example-based learning, improving from 0.016 (base) to 0.511 (one-shot) and further to 0.648 (CoT-1). This underscores Phi-4's notable in-context learning capabilities for schema lineage extraction tasks.
Comparatively, Qwen2.5-Coder-32B achieves performance on par with GPT-4o and GPT-4.1: its base prompting accuracy surpasses GPT-4o, while its one-shot and CoT-1 results are comparable to those of both proprietary LLMs.

\begin{figure}[t]
\centering
\begin{subfigure}[b]{0.48\textwidth}
    \includegraphics[width=\textwidth]{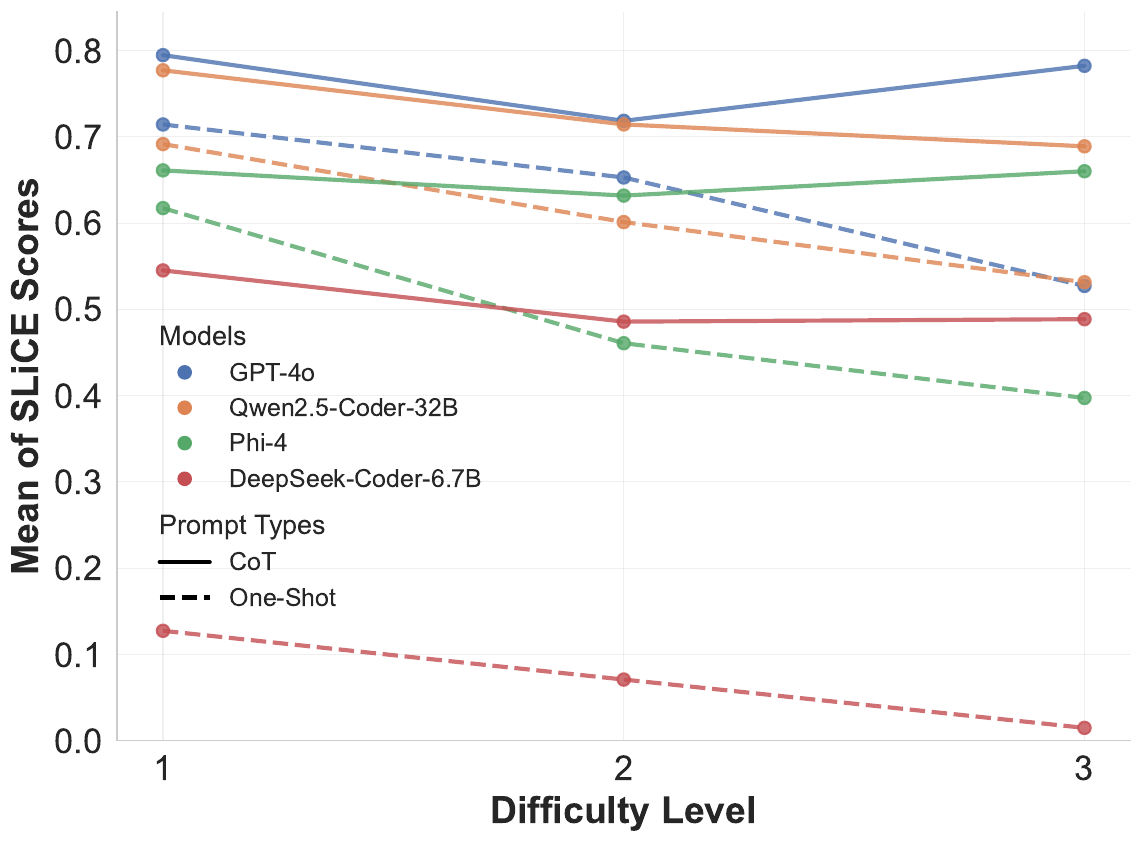}
    \caption{Average \slice{} scores across three script difficulty levels (1: easy, 2: medium, 3: hard) for four models under one-shot and CoT-1 prompting.}
    \label{fig:metrics-script-difficulty}
\end{subfigure}
\hfill
\begin{subfigure}[b]{0.48\textwidth}
    \includegraphics[width=\textwidth]{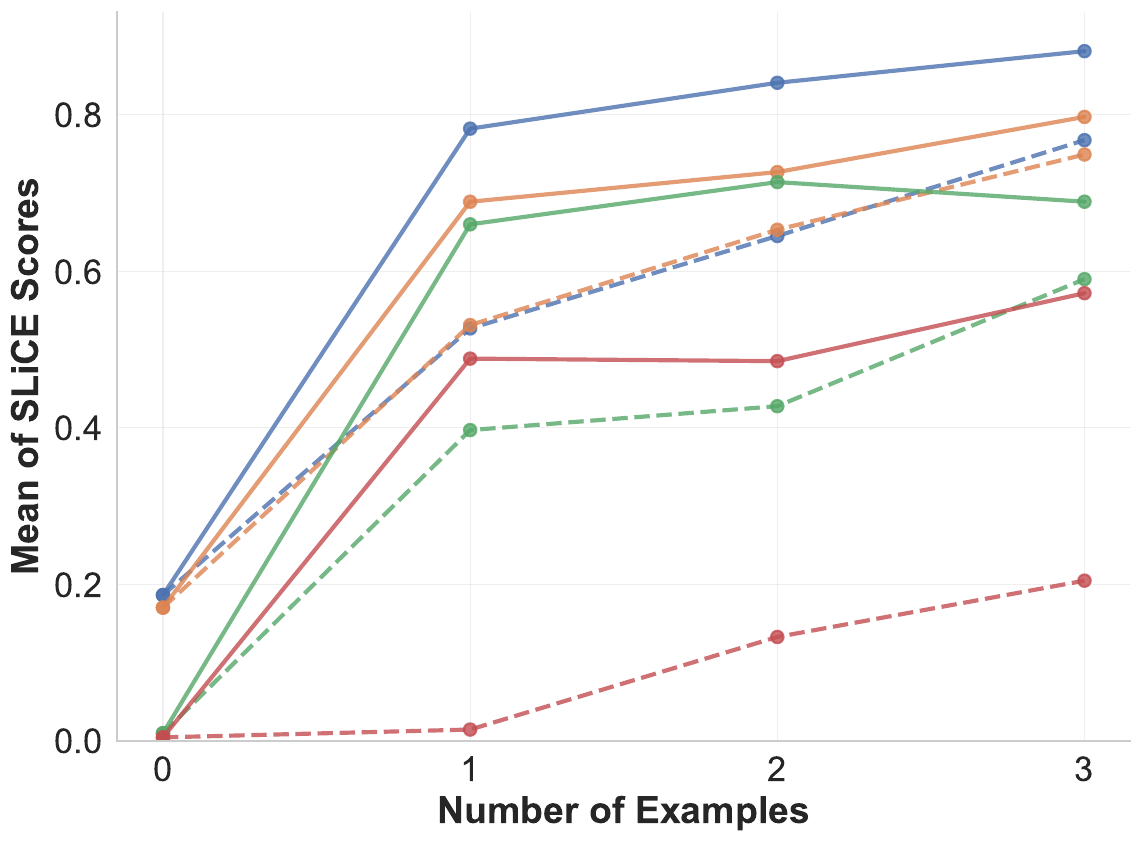}
    \caption{Average \slice{} scores on \textbf{hard} scripts with increasing numbers of examples (1–3) for few-shot and CoT prompting strategies.}
    \label{fig:metrics-model-prompt}
\end{subfigure}
\caption{Schema lineage extraction performance comparison across prompting strategies and script complexities for four models (GPT-4o, Qwen2.5-Coder-32B-Instruct, Phi-4-14B, and DeepSeek-Coder-6.7B). Line styles denote prompting strategies; colors indicate model variants. (a) shows the effect of script difficulty under different prompting strategies. (b) shows the effect of varying the number of examples in both few-shot and CoT prompting for hard scripts.}
\label{fig:overview}
\end{figure}

% We evaluate schema lineage extraction performance across three script difficulty levels, easy, medium, and hard, as detailed in Table~\ref{tab:dataset_stats}. 
To understand the impact of script complexity on extraction performance, we further stratify the \slice{} scores
% by script difficulty 
and illustrate the trends in Figure~\ref{fig:metrics-script-difficulty}. We select four representative models with varying scales: GPT-4o, Qwen2.5-Coder-32B, Phi-4, and DeepSeek-Coder-6.7B, using one-shot and CoT-1 prompting strategies across script difficulties. The rest of model performance is in Appendix~\ref{app-sec:addition-result}.

% Figure~\ref{fig:metrics-script-difficulty} shows a clear negative correlation between extraction performance and script complexity under the one-shot prompting scenario. 
Figure~\ref{fig:metrics-script-difficulty} reveals that schema lineage extraction performance decreases as script complexity increases across most scenarios which aligns with our design intuition.
When transitioning from one-shot to CoT-1 prompting, all models exhibit increased \slice{} scores, effectively mitigating the adverse effect of higher script complexity. This result underscores the significant benefit of incorporating even a single high-quality reasoning trace provided by a human expert into the prompt. For instance, Phi-4 (green color) achieves a \slice{} score of 0.660 on hard scripts using CoT-1 prompting (solid line), markedly surpassing the 0.397 score achieved with one-shot prompting (dash line). Additionally, the Qwen-2.5-Coder-32B under CoT-1 prompting (orange solid line) surpasses GPT-4o's performance under one-shot prompting (blue dash line) for scripts at all difficulty levels. This outcome is practically significant as it demonstrates that a 32B model, which can be internally deployed, can achieve performance comparable to the expensive GPT-4o.

We further investigate the effect of increasing the number of examples on schema lineage extraction performance, by analyzing average \slice{} scores for the hard scripts across the four representative models in Figure~\ref{fig:metrics-model-prompt}. 
%Given that complex scripts frequently generate extensive schema lineages with diverse format variations, additional schema lineage examples and corresponding human reasoning traces are introduced into the prompts.
Increasing the number of examples consistently enhances the \slice{} scores across all models, demonstrating a clear positive correlation between example quantity and performance improvement. CoT prompting generally outperforms few-shot prompting across all configurations. However, while CoT with 2-3 examples achieves superior performance, the magnitude of improvement remains modest. For instance, the Qwen2.5-Coder-32B model experiences a substantial increase of $23\%$ (from 0.531 to 0.653) from one-shot to two-shot prompting, whereas the improvement from CoT-1 to CoT-2 is considerably smaller at only $6\%$ (from 0.689 to 0.727). This pattern suggests that schema lineage extraction benefits substantially from a single high-quality reasoning trace, with additional reasoning traces yielding diminishing returns.

\section{Discussion}
% 1. key findings in experiments 2. rag application example 3. score for finetuning.

Our experiments demonstrate that the proposed \slice{} metric effectively captures schema lineage extraction performance across varying language models and prompting strategies. While proprietary LLMs deliver strong extraction performance, each prompt must contain complete data pipeline scripts, which can exceed hundreds of thousands of tokens, leading to cost escalation. Our work reveals that open-source models at the 32B parameter scale, when augmented with chain-of-thought reasoning traces, achieve extraction performance comparable to proprietary state-of-the-art LLMs such as GPT-4o and GPT-4.1. Incorporating even a single high-quality reasoning trace remarkably enhances performance. However, the requirement for human experts to provide reasoning trace examples for each script type limits scalability.

A primary application enabled by accurate schema lineage extraction is the automated creation of  high-quality documentation alongside dynamic data pipeline scripts. This documentation subsequently serves as a robust knowledge base for RAG systems. Take the schema lineage extraction in Figure \ref{fig:schema-lineage} as an example. The schema \texttt{TotalAmountSpent} originates from the database columns \texttt{customer\_id} and \texttt{amount}, with their definitions sourced from the database's metadata. Schema lineage explicitly traces transformations and aggregations, empowering the LLM to generate a precise and contextual business statement: \textit{"TotalAmountSpent shows the total amount spent by each customer by aggregating individual transaction amounts. ...<business impact provided by LLM knowledge>"}.
Such detailed, dynamic, and domain-specific documentation significantly enriches downstream AI applications. Furthermore, accurate schema lineage substantially improves text-to-SQL tasks by providing precise definitions and relevant business contexts, ultimately enhancing AI-driven analytical workflows from human queries.

\section{Conclusion}
In this paper, we proposed an innovative framework for automated schema lineage extraction tailored to multi-language enterprise data pipelines. Recognizing the inherent semantic drift due to complex data transformations, our approach systematically captures schema lineage details (source schemas, tables, transformation logic, and aggregation operations) directly from pipeline scripts. We curated a robust benchmark dataset consisting of 1,700 schema annotations stratified across varying script complexities, representative of real-world industry scenarios. Central to our methodology is the \slice{} score, a composite evaluation metric that combines structural correctness with semantic precision. This metric enables granular diagnosis for the lineage of real-world applications.
mportantly, provides a well-structured reward signal that can be leveraged for fine-tuning language models in future work, offering a direct path toward improving model alignment with schema lineage extraction tasks.

Our experimental analysis examined multiple state-of-the-art language models under diverse prompting strategies. Key findings revealed that the performance of schema lineage extraction significantly improves with increasing model size and contextual richness in prompts. Specifically, chain-of-thought reasoning significantly enhance extraction performance. We observed that 32B SLM achieves performance levels comparable to proprietary LLMs, highlighting their viability for enterprise deployment.

The proposed method directly facilitates high-quality dynamic documentation, significantly enhancing downstream applications such as RAG and text-to-SQL systems. By providing accurate, contextually-rich schema documentation, our approach empowers enterprises to maintain rigorous data governance and analytical reproducibility, effectively bridging the semantic gap in enterprise data transformation processes.

\begin{ack}
% Acknowledgments go here. Remember to declare funding and competing interests.
% Do not include this section in the anonymized submission, only in the final paper.
This work is supported by our manager Cheng Wu. We thank Lili Che and Naga Sai Kiran Kambhampati for curating the high-quality data pipeline scripts and annotating the schema lineage.
\end{ack}

% IEEE citation style: if number of authors is more than 6, only keep the first six authors and use "and others" in the bibtex
\bibliographystyle{unsrtnat}
\bibliography{references}

\newpage
%%%%%%%%%%%%%%%%%%%%%%%%%%%%%%%%%%%%%%%%%%%%%%%%%%%%%%%%%%%%

\appendix

\section {Data Gallery}
\renewcommand{\thefigure}{A.\arabic{figure}}
\setcounter{figure}{0}
\renewcommand{\thetable}{A.\arabic{table}}
\setcounter{table}{0}

% data script; how to define easy, medium, hard; prompt structure
% move figure 3 to here
\subsection{Script Difficulty}
\label{sec:script-complexity}
To quantify the complexity of data pipeline scripts, we introduce a scoring framework that evaluates each script based on its structural and operational features. Scripts are scored from 0 to 3 across three independent dimensions: data sources, transformations, and aggregations. A point is awarded for each dimension that demonstrates a higher level of complexity, as detailed below:

\begin{itemize}
\item \textbf{Data Sources}:
\begin{itemize}
\item 0 points for scripts accessing one or two distinct data sources.
\item +1 point for scripts accessing three or more distinct data sources.
\end{itemize}

\item \textbf{Transformation}:
\begin{itemize}
    \item 0 points for scripts with only basic transformations (e.g., column renaming, type casting).
    \item +1 point for scripts that include a transformation chain, where the output of one operation serves as the input to another.
\end{itemize}

\item \textbf{Aggregation}:
\begin{itemize}
    \item 0 points for scripts with no aggregation or pivot operations.
    \item +1 point for scripts containing any aggregation function (e.g., \texttt{SUM}, \texttt{COUNT}, \texttt{PIVOT}).
\end{itemize}
\end{itemize}

The total complexity score for a script is the sum of the points from each dimension:
\begin{equation*}
\text{Total Score} = \text{Points(Data Sources)} + \text{Points(Transformation)} + \text{Points(Aggregation)}
\end{equation*}

The final score determines the script's difficulty level, as defined in Table~\ref{tab:script-difficulty}.

\begin{table}[ht]
\centering
\caption{Difficulty levels for script data based on scoring criteria.}
\label{tab:script-difficulty}
\small
\begin{tabular}{@{}>{\centering\arraybackslash}m{3cm} >{\centering\arraybackslash}m{2cm} m{7cm}@{}}
\toprule
\textbf{Difficulty Level} & \textbf{Score} & \textbf{Description} \\
\midrule
Level 1: Easy & 0 or 1 & Scripts with minimal complexity, exhibiting at most one complexity factor (e.g., multiple data sources, a transformation chain, or an aggregation). \\
\hline
Level 2: Medium & 2 & Scripts incorporating two of the three complexity factors, such as multiple sources with a transformation chain but no aggregation. \\
\hline
Level 3: Hard & 3 & Scripts featuring all three complexity factors: multiple data sources (
geq3), chained transformations, and at least one aggregation or pivot operation. \\
\bottomrule\\
\end{tabular}
\end{table}

\subsection{Scripts}
\label{section: script}
To trace schema lineage from real-world scripts that frequently incorporate multiple programming languages, we developed a custom parsing strategy capable of handling multi-language code environments. Modern data processing workflows typically employ different programming languages optimized for specific computational tasks. Data engineers commonly utilize Python with specialized libraries such as PySpark, an interface for Apache Spark that enables distributed processing of large datasets across cluster computing environments. This approach facilitates efficient large-scale data cleaning and transformation operations. Subsequently, analysts and business users employ SQL for analytics and reporting tasks on the processed data. Listing \ref{lst:example} presents a synthetic script demonstrating the integration of Python and SQL components with level of difficulty as hard. We employ the delimiter \lstinline[basicstyle=\ttfamily]{>>>>>} to denote programming language transitions during the parsing process.
Our schema lineage tracing algorithm operates using a bottom-up traversal approach, initiating from a pre-specified target column and propagating upward through the computational graph. All transformation and aggregation operations that influence the target column are captured and recorded according to our standardized schema lineage representation format. 
One complete example of schema lineage is showed in Table~\ref{tab:lineage_example}.

\lstdefinelanguage{PythonWithSQL}{
  language=Python,
  morekeywords={ SELECT,FROM,WHERE,INSERT,INTO,UPDATE,SET,DELETE,JOIN,ON,AS,AND,OR,NOT,NULL,LIMIT,ORDER,BY,GROUP
  },
  keywordstyle=\color{blue},
  commentstyle=\color{gray},
  stringstyle=\color{red},
  showstringspaces=false,
  columns=flexible,
  basicstyle=\ttfamily\small,
  numbers=left,
  numberstyle=\tiny\color{gray},
  breaklines=true,
  frame=single,
  tabsize=2,
}

\begin{lstlisting}[language=PythonWithSQL, caption={Python Code + SQL}, label=lst:example]
import pyspark.sql.functions as F
from pyspark.sql import SparkSession
from pyspark.sql.types import StructType, StructField, IntegerType, StringType, DateType, TimestampType, DecimalType, DoubleType
from datetime import datetime, timedelta

def clean_customers(df):
    email_cleaned = df.withColumn(
        "email_address",
        F.when(
            F.col("email_address").isNull() | 
            F.lower(F.col("email_address")).isin("", "invalid-email", "none"),
            F.lit("invalid_format@example.com")
        ).otherwise(F.col("email_address"))
    )

    names_formatted = email_cleaned \
        .withColumn("first_name", F.initcap("first_name")) \
        .withColumn("last_name", F.initcap("last_name"))

    gender_normalized = names_formatted.withColumn(
        "gender",
        F.when(
            F.lower("gender").isin("none", "prefer not to say"),
            F.lit("Prefer Not To Say")
        ).otherwise(F.initcap("gender"))
    )

    location_normalized = gender_normalized.withColumn(
        "state_province", F.upper("state_province")
    )

    phone_cleaned = location_normalized.withColumn(
        "phone_number", F.regexp_replace("phone_number", "[^0-9]", "")
    )

    registration_parsed = phone_cleaned.withColumn(
        "registration_date", F.to_timestamp("registration_date")
    ).filter(F.col("registration_date").isNotNull())

    premium_flagged = registration_parsed.withColumn(
        "is_premium_member",
        F.when(F.lower("is_premium_member").isin("true", "1", "yes"), F.lit(True))
         .otherwise(F.lit(False))
    )

    deduplicated = premium_flagged.dropDuplicates(["customer_id"])

    return deduplicated

def clean_accounts(df):
    balance_casted = df.withColumn("balance", F.col("balance").cast(DecimalType(18, 2)))

    balance_cleaned = balance_casted.withColumn(
        "balance",
        F.when(F.col("balance").isNull() | (F.col("balance") < 0), F.lit(0.00))
        .otherwise(F.col("balance"))
    )

    account_type_cleaned = balance_cleaned.withColumn(
        "account_type",
        F.when(F.lower("account_type").isin("none", "unspecified"), F.lit("Unspecified"))
         .otherwise(F.initcap("account_type"))
    )

    status_formatted = account_type_cleaned.withColumn("status", F.initcap("status"))

    opening_date_parsed = status_formatted.withColumn(
        "opening_date", F.to_date("opening_date")
    ).filter(F.col("opening_date").isNotNull())

    interest_casted = opening_date_parsed.withColumn(
        "interest_rate", F.col("interest_rate").cast(DoubleType())
    )

    credit_limit_casted = interest_casted.withColumn(
        "credit_limit", F.col("credit_limit").cast(DecimalType(18, 2))
    )

    return credit_limit_casted.dropDuplicates(["account_id"])


def clean_transactions(df):
    df = df.withColumn("transaction_timestamp", F.to_timestamp("transaction_timestamp")) \
           .filter(F.col("transaction_timestamp").isNotNull()) \
           .withColumn(
               "transaction_timestamp",
               F.when(F.col("transaction_timestamp") > F.current_timestamp(), F.current_timestamp())
                .otherwise(F.col("transaction_timestamp"))
           ) \
           .withColumn("amount", F.col("amount").cast(DecimalType(18, 2))) \
           .withColumn("amount", F.when(F.col("amount").isNull(), F.lit(0.00)).otherwise(F.abs("amount"))) \
           .withColumn(
               "transaction_type",
               F.when(F.lower("transaction_type").isin("unknown", "none"), F.lit("Other"))
                .otherwise(F.initcap("transaction_type"))
           ) \
           .withColumn("status", F.initcap("status"))

    return df.dropDuplicates()

if __name__ == "__main__":
    spark.conf.set("spark.storage.synapse.linkedServiceName",linked_service_name)
    spark.conf.set("fs.azure.account.oauth.provider.type","com.bank.azure.synapse.tokenlibrary.LinkedServiceBasedTokenProvider")

    raw_customers_df = spark.read.load('abfss://bank@efgh.dfs.core.windows.net/raw_customers/customers.parquet',format='parquet')
    raw_accounts_df = spark.read.load('abfss://bank@efgh.dfs.core.windows.net/raw_accounts/accounts.parquet',format='parquet')
    raw_transactions_df = spark.read.load('abfss://bank@efgh.dfs.core.windows.net/raw_transactions/transactions.parquet',format='parquet')

    print("\n--- Applying Cleaning Transformations ---")
    cleaned_customers_df = clean_customers(raw_customers_df)
    cleaned_accounts_df = clean_accounts(raw_accounts_df)
    cleaned_transactions_df = clean_transactions(raw_transactions_df)

    OutputPath_customers='abfss://bank@efgh.dfs.core.windows.net/customersprod/customers.parquet'
    OutputPath_accounts='abfss://bank@efgh.dfs.core.windows.net/accountprod/accounts.parquet'
    OutputPath_transactions='abfss://bank@efgh.dfs.core.windows.net/transactionsprod/transactions.parquet'

    cleaned_customers_df.write.mode('overwrite').parquet(OutputPath_customers)
    cleaned_accounts_df.write.mode('overwrite').parquet(OutputPath_accounts)
    cleaned_transactions_df.write.mode('overwrite').parquet(OutputPath_transactions)


>>>>>

SELECT
    C.customer_id AS CustomerId,
    C.first_name AS FirstName,
    C.last_name AS LastName,
    C.is_premium_member AS IsPremiumMember,
    C.registration_date AS CustomerRegistrationDate,
    A.account_type AS AccountType,
    A.balance AS CurrentAccountBalance,
    A.credit_limit AS AccountCreditLimit,
    A.opening_date AS AccountOpeningDate,
    SUM(T.amount) AS TotalAmountSpent,
    COUNT(T.transaction_id) AS MonthlyTransactionCount,
    AVG(T.amount) AS AverageMonthlyTransactionAmount
FROM
    Customers AS C
INNER JOIN
    Accounts AS A ON C.customer_id = A.customer_id
INNER JOIN
    Transactions AS T ON A.account_id = T.account_id
WHERE
    T.transaction_timestamp >= '2025-05-01' AND T.transaction_timestamp < '2025-06-01'
    AND T.transaction_type IN ('Withdrawal', 'Purchase', 'Bill Payment', 'Transfer-Out')
    AND T.status = 'Completed'
GROUP BY
    C.customer_id,
    C.first_name,
    C.last_name,
    C.is_premium_member,
    C.registration_date,
    A.account_type,
    A.balance,
    A.credit_limit,
    A.opening_date
ORDER BY
    TotalAmountSpent DESC, C.customer_id, A.account_type;

\end{lstlisting}

\lstset{
  basicstyle=\ttfamily\small,
  breaklines=true, 
  columns=fullflexible, 
  breakatwhitespace=false,
  breakindent=0pt,
  showstringspaces=false,
  frame=none
}

\begin{table}[htbp]
\centering
\caption{Schema Lineage for the \texttt{AverageMonthlyTransactionAmount} column from Listing~\ref{lst:example}}
\label{tab:lineage_example}
\begin{tabular}{>{\centering\arraybackslash}p{2.5cm}|>{\centering\arraybackslash}m{12cm}}
\hline
\textbf{Key} & \textbf{Value} \\
\hline
\texttt{source\_schema} &

\begin{lstlisting}
amount, customer_id, first_name, last_name, is_premium_member, registration_date, account_type, balance, credit_limit, opening_date, transaction_timestamp, transaction_type, status
\end{lstlisting}
\\[-1.5ex]
\hline
\texttt{source\_table} &
\begin{lstlisting}
abfss://bank@efgh.dfs.core.windows.net/raw\_customers/customers.parquet;
abfss://bank@efgh.dfs.core.windows.net/raw\_accounts/accounts.parquet;
abfss://bank@efgh.dfs.core.windows.net/raw\_transactions/transactions.parquet
\end{lstlisting}

\\[-1.5ex]
\hline
\texttt{transformation} &
\begin{lstlisting}
C.customer_id AS CustomerId <CODEEND> email_cleaned.withColumn("first_name", F.initcap("first_name")) <CODEEND> C.first_name AS FirstName <CODEEND> email_cleaned.withColumn("last_name", F.initcap("last_name")) <CODEEND> C.last_name AS LastName <CODEEND> registration_parsed.withColumn("is_premium_member", F.when(F.lower("is_premium_member").isin("true", "1", "yes"), F.lit(True)).otherwise(F.lit(False))) <CODEEND> C.is_premium_member AS IsPremiumMember <CODEEND> phone_cleaned.withColumn("registration_date", F.to_timestamp("registration_date")) <CODEEND> C.registration_date AS CustomerRegistrationDate <CODEEND> balance_cleaned.withColumn("account_type", F.when(F.lower("account_type").isin("none", "unspecified"), F.lit("Unspecified")).otherwise(F.initcap("account_type"))) <CODEEND> A.account_type AS AccountType <CODEEND> df.withColumn("balance", F.col("balance").cast(DecimalType(18, 2))) <CODEEND> balance_casted.withColumn("balance", F.when(F.col("balance").isNull() | (F.col("balance") < 0), F.lit(0.00)).otherwise(F.col("balance"))) <CODEEND> A.balance AS CurrentAccountBalance <CODEEND> interest_casted.withColumn("credit_limit", F.col("credit_limit").cast(DecimalType(18, 2))) <CODEEND> A.credit_limit AS AccountCreditLimit <CODEEND> status_formatted.withColumn("opening_date", F.to_date("opening_date")) <CODEEND> A.opening_date AS AccountOpeningDate <CODEEND> df.withColumn("amount", F.when(F.col("amount").isNull(), F.lit(0.00)).otherwise(F.abs("amount"))) <CODEEND> df.withColumn("amount", F.col("amount").cast(DecimalType(18, 2))) <CODEEND> AVG(T.amount) AS AverageMonthlyTransactionAmount
\end{lstlisting}

\\[-1.5ex]
\hline
\texttt{aggregation} &
\begin{lstlisting}
AVG() GROUP BY C.customer_id, C.first_name, C.last_name, C.is_premium_member, C.registration_date, A.account_type, A.balance, A.credit_limit, A.opening_date
\end{lstlisting}
\\
\hline

\end{tabular}
\end{table}

\newpage

\subsection{Prompts}\label{app-sec:prompts}
We present the prompt templates designed for our schema lineage extraction task, structured around three distinct prompting strategies: \textbf{Base}, \textbf{Few-Shot}, and \textbf{Chain-of-Thought (CoT)}. Each template incorporates placeholders for the data pipeline script, with examples included exclusively in the Few-Shot and CoT configurations. The number of examples provided scales according to input script complexity: one example for \textit{Easy} cases, up to two for \textit{Medium} complexity, and up to three for \textit{Hard} scenarios. All prompt templates direct the model to generate structured output conforming to a specified JSON-like schema enclosed within \texttt{<answer>} \texttt{</answer>} tags. The CoT template uniquely incorporates an intermediate reasoning step delimited by \texttt{<think>} \texttt{</think>} tags to facilitate explicit reasoning processes.
\begin{figure}[H]
\centering
\includegraphics[width=0.55\textwidth]{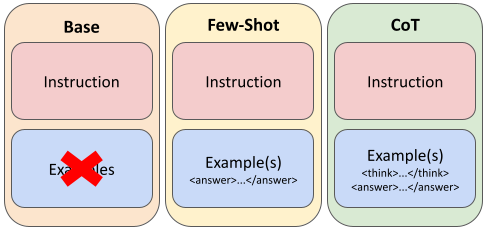}
\caption{Comparison of prompting strategies for schema lineage extraction. Base prompting provides only task instructions, few-shot prompting incorporates example outputs to demonstrate the expected format, and Chain-of-Thought (CoT) prompting includes explicit reasoning traces that guide the lineage process step-by-step.}
\label{fig:prompt-stratagy}
\end{figure}

\lstset{
  basicstyle=\ttfamily\small,
  breaklines=true,
  columns=fullflexible
}

\textbf{Base Prompt Template}
\begin{lstlisting}[language=Python]
You are a data lineage analysis assistant. Your task is to analyze the provided data generation script and trace the lineage of a specific column which is specified by the user.

Your response must include <answer> </answer> part:
<answer> {
  "source_schema": "...",
  "source_table": "...",
  "transformation": "...",
  "aggregation": "..."
} </answer>.

... (additional instructions omitted for brevity) ...

Data Pipeline Script: YOUR PINELINE SCRIPT
\end{lstlisting}

\textbf{Few-Shot Prompt Template (1, 2, or 3 examples)}
\begin{lstlisting}[language=Python]
You are a data lineage analysis assistant. Your task is to analyze the provided data generation script and trace the lineage of a specific column which is specified by the user.

Your response must include <answer> </answer> part:
<answer> {
  "source_schema": "...",
  "source_table": "...",
  "transformation": "...",
  "aggregation": "..."
} </answer>.

... (additional instructions omitted for brevity) ...

Data Pipeline Script: YOUR PIPELINE SCRIPT

Examples: YOUR OUTPUT EXAMPLE(S)
"""
\end{lstlisting}

\textbf{Chain-of-Thought Prompt Template (1, 2, or 3 examples)}
\begin{lstlisting}[language=Python]
You are a data lineage analysis assistant. Your task is to analyze the provided data generation script and trace the lineage of a specific column which is specified by the user.

Your response must include two parts:
1. <think> ... </think>
2. <answer> {{
  "source_schema": "...",
  "source_table": "...",
  "transformation": "...",
  "aggregation": "..."
}} </answer>.

... (additional instructions omitted for brevity) ...

Data Pipeline Script: YOUR PIPELINE SCRIPT

Examples: YOUR OUTPUT EXAMPLE(S)
"""
\end{lstlisting}

\newpage
\section{Additional Results}
\label{app-sec:addition-result}

\renewcommand{\thefigure}{B.\arabic{figure}}
\setcounter{figure}{0}
\renewcommand{\thetable}{B.\arabic{figure}}
\setcounter{table}{0}
% Additional experimental results, if any
\begin{figure}[htbp]
\centering
\includegraphics[width=0.8\textwidth]{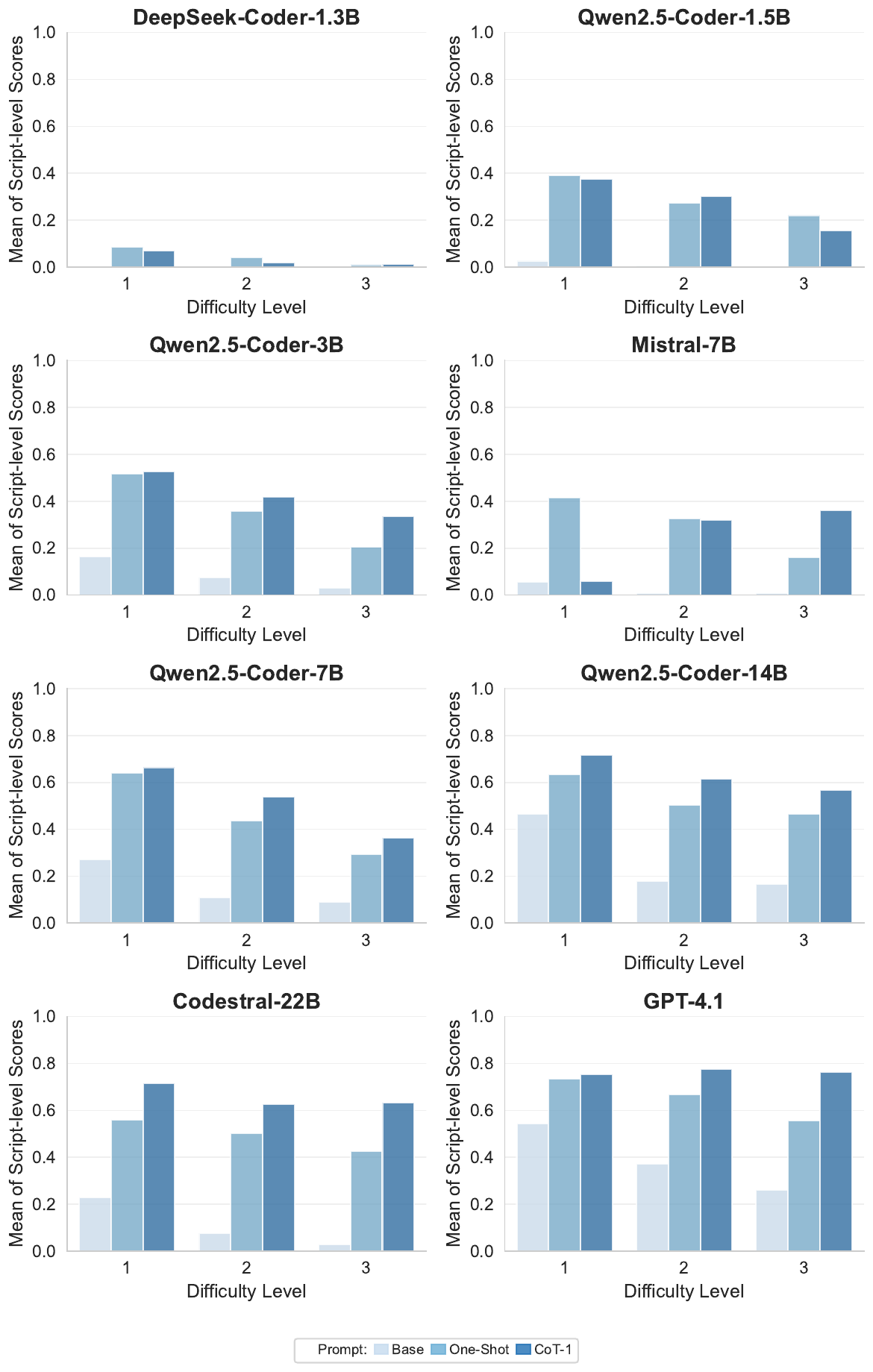}
\caption{\textbf{Performance comparison across models and prompt strategies by 
  script difficulty}. Bar plots show mean script-level scores for eight language
   models across three prompting strategies: base (no
  additional output examples), few-shot (one example), and CoT (one reasoning trace exmaple). Scripts are grouped by difficulty level (1-3), with higher difficulty indicating more complex reasoning requirements. Larger models consistently outperform smaller ones. CoT prompting
  provides moderate improvements across most difficulty levels.}
\label{fig:all-model-performance}
\end{figure}

\begin{figure}[htbp]
    \centering
    \includegraphics[width=0.7\columnwidth]{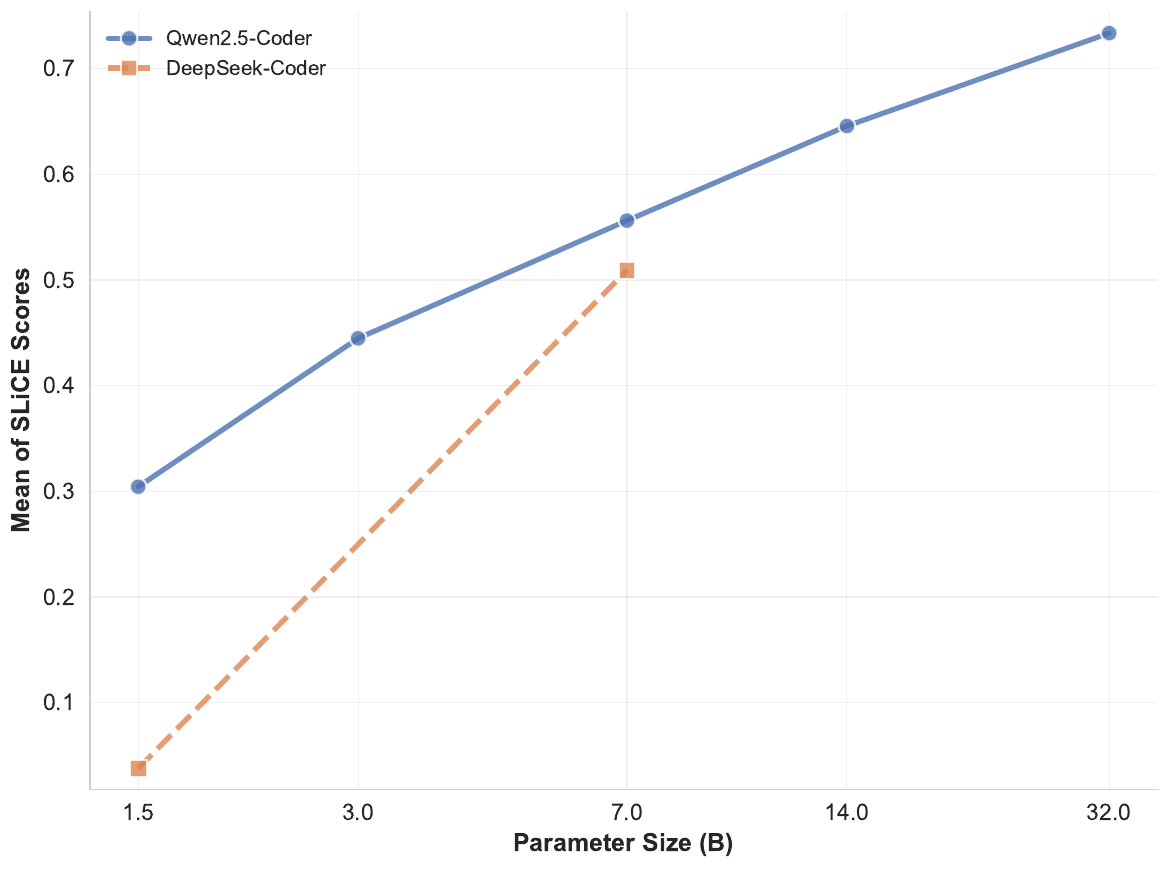}
    \caption{\textbf{Model performance scaling with parameter size for Chain-of-Thought prompting}. The plot shows how mean \slice{} vary with model parameter size (in billions) for Qwen2.5-Coder (1.5B, 3B, 7B, 14B, 32B) and DeepSeek-Coder (1.3B, 6.7B) model families when having one human reasoning trace in the prompt. Both model families demonstrate improved performance with increased parameter size, with Qwen2.5-Coder models consistently outperforming DeepSeek-Coder models across all parameter scales.}
    \label{fig:model_performance_vs_param_size}
\end{figure}

\begin{figure}[htbp]
\centering
\includegraphics[width=0.7\textwidth]{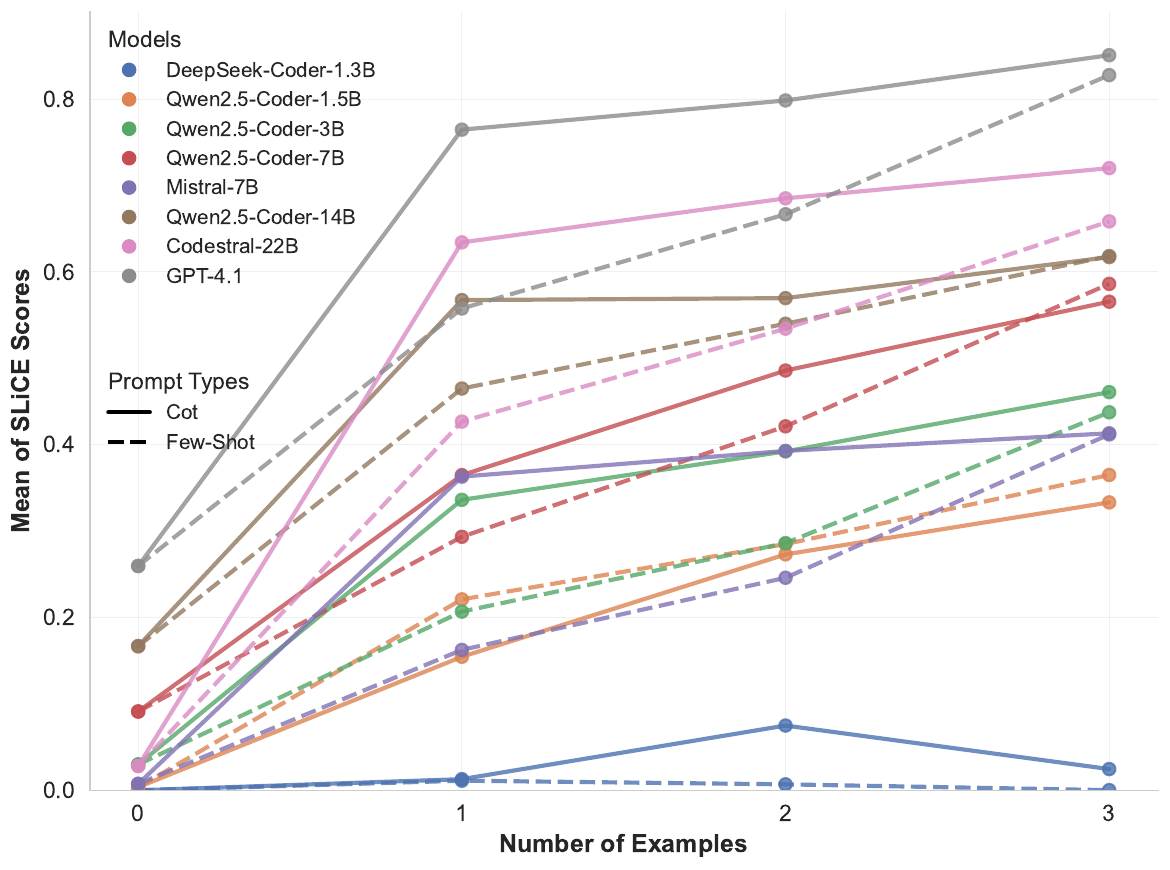}
\caption{
\textbf{Average \slice{}  across language models and prompting strategies for hard scripts}. Y-axis the average \slice{} for 8 language models (GPT-4.1, Codestral-22B, Qwen2.5-Coder (7B, 3B, 1.5B), Mistral-7B, and DeepSeek-Coder-1.5B) across different prompting strategies: base (no additional output examples), few-shot (one example), and CoT (one reasoning trace example). The results indicate that larger models generally perform better, with CoT prompting yielding higher metrics than few-shot prompting.}
\label{fig:metrics-model-prompt-others}
\end{figure}

\newpage

\begin{table}[htbp]
\centering
\caption{Benchmark results of language models on schema lineage extraction (Part 1): Qwen2.5-Coder variants. Mean corpus-level \slice{} scores and standard deviations across six random seeds (Mean $\pm$ Standard Deviation).}
\label{tab:overall_performance_part1}
\small
\begin{tabular}{l|l|ccccc}
\toprule
Parameter & Difficulty & \multicolumn{5}{c}{Qwen2.5-Coder} \\
\cmidrule(lr){3-7}
 & & 1.5B & 3B & 7B & 14B & 32B \\
\midrule
\multirow{3}{*}{Base} & Easy & 0.027 $\pm$ 0.004 & 0.163 $\pm$ 0.004 & 0.271 $\pm$ 0.011 & 0.465 $\pm$ 0.010 & 0.578 $\pm$ 0.007 \\
 & Medium & 0.006 $\pm$ 0.002 & 0.074 $\pm$ 0.005 & 0.108 $\pm$ 0.006 & 0.179 $\pm$ 0.006 & 0.239 $\pm$ 0.004 \\
 & Hard & 0.004 $\pm$ 0.001 & 0.030 $\pm$ 0.006 & 0.091 $\pm$ 0.005 & 0.167 $\pm$ 0.009 & 0.171 $\pm$ 0.006 \\
\cmidrule(lr){1-7}
\multirow{3}{*}{One-shot} & Easy & 0.391 $\pm$ 0.009 & 0.517 $\pm$ 0.017 & 0.639 $\pm$ 0.013 & 0.635 $\pm$ 0.006 & 0.692 $\pm$ 0.004 \\
 & Medium & 0.274 $\pm$ 0.012 & 0.358 $\pm$ 0.021 & 0.436 $\pm$ 0.031 & 0.504 $\pm$ 0.009 & 0.601 $\pm$ 0.007 \\
 & Hard & 0.221 $\pm$ 0.006 & 0.207 $\pm$ 0.008 & 0.293 $\pm$ 0.029 & 0.465 $\pm$ 0.024 & 0.531 $\pm$ 0.010 \\
\cmidrule(lr){1-7}
\multirow{2}{*}{Two-shot} & Medium & 0.345 $\pm$ 0.007 & 0.467 $\pm$ 0.012 & 0.547 $\pm$ 0.020 & 0.586 $\pm$ 0.008 & 0.664 $\pm$ 0.009 \\
 & Hard & 0.285 $\pm$ 0.030 & 0.286 $\pm$ 0.025 & 0.421 $\pm$ 0.018 & 0.540 $\pm$ 0.009 & 0.653 $\pm$ 0.007 \\
\cmidrule(lr){1-7}
\multirow{1}{*}{Three-shot} & Hard & 0.365 $\pm$ 0.026 & 0.438 $\pm$ 0.027 & 0.586 $\pm$ 0.052 & 0.618 $\pm$ 0.016 & 0.749 $\pm$ 0.015 \\
\cmidrule(lr){1-7}
\multirow{3}{*}{CoT-1} & Easy & 0.377 $\pm$ 0.021 & 0.528 $\pm$ 0.003 & 0.664 $\pm$ 0.014 & 0.717 $\pm$ 0.014 & 0.777 $\pm$ 0.010 \\
 & Medium & 0.302 $\pm$ 0.014 & 0.418 $\pm$ 0.021 & 0.540 $\pm$ 0.013 & 0.616 $\pm$ 0.014 & 0.714 $\pm$ 0.010 \\
 & Hard & 0.154 $\pm$ 0.036 & 0.336 $\pm$ 0.009 & 0.365 $\pm$ 0.019 & 0.567 $\pm$ 0.012 & 0.689 $\pm$ 0.019 \\
\cmidrule(lr){1-7}
\multirow{2}{*}{CoT-2} & Medium & 0.293 $\pm$ 0.015 & 0.437 $\pm$ 0.011 & 0.568 $\pm$ 0.015 & 0.685 $\pm$ 0.017 & 0.748 $\pm$ 0.014 \\
 & Hard & 0.273 $\pm$ 0.013 & 0.392 $\pm$ 0.011 & 0.486 $\pm$ 0.015 & 0.570 $\pm$ 0.012 & 0.727 $\pm$ 0.024 \\
\cmidrule(lr){1-7}
\multirow{1}{*}{CoT-3} & Hard & 0.333 $\pm$ 0.010 & 0.461 $\pm$ 0.012 & 0.566 $\pm$ 0.029 & 0.617 $\pm$ 0.014 & 0.797 $\pm$ 0.019 \\
\bottomrule
\end{tabular}
\end{table}

\begin{table}[htbp]
\centering
\caption{Benchmark results of language models on schema lineage extraction (Part 2): DeepSeek-Coder models. Mean corpus-level \slice{} scores and standard deviations across six random seeds (Mean $\pm$ Standard Deviation).}
\label{tab:overall_performance_part2}
\small
\begin{tabular}{l|l|cc}
\toprule
Parameter & Difficulty & \multicolumn{2}{c}{DeepSeek-Coder} \\
\cmidrule(lr){3-4}
 & & 1.3B & 6.7B \\
\midrule
\multirow{3}{*}{Base} & Easy & 0.000 $\pm$ 0.000 & 0.006 $\pm$ 0.004 \\
 & Medium & 0.000 $\pm$ 0.000 & 0.000 $\pm$ 0.001 \\
 & Hard & 0.000 $\pm$ 0.000 & 0.005 $\pm$ 0.007 \\
\cmidrule(lr){1-4}
\multirow{3}{*}{One-shot} & Easy & 0.085 $\pm$ 0.023 & 0.127 $\pm$ 0.024 \\
 & Medium & 0.042 $\pm$ 0.016 & 0.071 $\pm$ 0.025 \\
 & Hard & 0.011 $\pm$ 0.008 & 0.015 $\pm$ 0.011 \\
\cmidrule(lr){1-4}
\multirow{2}{*}{Two-shot} & Medium & 0.007 $\pm$ 0.006 & 0.143 $\pm$ 0.044 \\
 & Hard & 0.007 $\pm$ 0.005 & 0.133 $\pm$ 0.037 \\
\cmidrule(lr){1-4}
\multirow{1}{*}{Three-shot} & Hard & 0.000 $\pm$ 0.001 & 0.205 $\pm$ 0.051 \\
\cmidrule(lr){1-4}
\multirow{3}{*}{CoT-1} & Easy & 0.070 $\pm$ 0.016 & 0.545 $\pm$ 0.010 \\
 & Medium & 0.019 $\pm$ 0.019 & 0.486 $\pm$ 0.013 \\
 & Hard & 0.013 $\pm$ 0.021 & 0.489 $\pm$ 0.018 \\
\cmidrule(lr){1-4}
\multirow{2}{*}{CoT-2} & Medium & 0.043 $\pm$ 0.028 & 0.501 $\pm$ 0.009 \\
 & Hard & 0.075 $\pm$ 0.019 & 0.485 $\pm$ 0.008 \\
\cmidrule(lr){1-4}
\multirow{1}{*}{CoT-3} & Hard & 0.025 $\pm$ 0.009 & 0.573 $\pm$ 0.023 \\
\bottomrule
\end{tabular}
\end{table}

\begin{table}[htbp]
\centering
\caption{Benchmark results of language models on schema lineage extraction (Part 3): GPT models and other language models. Mean corpus-level \slice{} scores and standard deviations across six random seeds (Mean $\pm$ Standard Deviation).}
\label{tab:overall_performance_part3}
\small
\begin{tabular}{l|l|ccccc}
\toprule
Parameter & Difficulty & GPT-4.1 & GPT-4o & Phi-4 & Codestral-22B & Mistral-7B \\
\midrule
\multirow{3}{*}{Base} & Easy & 0.544 $\pm$ 0.006 & 0.379 $\pm$ 0.005 & 0.017 $\pm$ 0.005 & 0.230 $\pm$ 0.005 & 0.057 $\pm$ 0.004 \\
 & Medium & 0.373 $\pm$ 0.007 & 0.241 $\pm$ 0.004 & 0.018 $\pm$ 0.005 & 0.077 $\pm$ 0.009 & 0.008 $\pm$ 0.003 \\
 & Hard & 0.260 $\pm$ 0.007 & 0.186 $\pm$ 0.006 & 0.010 $\pm$ 0.005 & 0.028 $\pm$ 0.003 & 0.007 $\pm$ 0.001 \\
\cmidrule(lr){1-7}
\multirow{3}{*}{One-shot} & Easy & 0.734 $\pm$ 0.004 & 0.714 $\pm$ 0.003 & 0.617 $\pm$ 0.007 & 0.561 $\pm$ 0.005 & 0.416 $\pm$ 0.007 \\
 & Medium & 0.668 $\pm$ 0.016 & 0.653 $\pm$ 0.017 & 0.461 $\pm$ 0.006 & 0.503 $\pm$ 0.007 & 0.327 $\pm$ 0.009 \\
 & Hard & 0.558 $\pm$ 0.017 & 0.527 $\pm$ 0.007 & 0.397 $\pm$ 0.012 & 0.427 $\pm$ 0.003 & 0.163 $\pm$ 0.005 \\
\cmidrule(lr){1-7}
\multirow{2}{*}{Two-shot} & Medium & 0.751 $\pm$ 0.008 & 0.685 $\pm$ 0.010 & 0.548 $\pm$ 0.011 & 0.571 $\pm$ 0.008 & 0.427 $\pm$ 0.008 \\
 & Hard & 0.667 $\pm$ 0.007 & 0.645 $\pm$ 0.009 & 0.428 $\pm$ 0.038 & 0.534 $\pm$ 0.009 & 0.246 $\pm$ 0.012 \\
\cmidrule(lr){1-7}
\multirow{1}{*}{Three-shot} & Hard & 0.828 $\pm$ 0.008 & 0.768 $\pm$ 0.010 & 0.590 $\pm$ 0.029 & 0.658 $\pm$ 0.010 & 0.412 $\pm$ 0.010 \\
\cmidrule(lr){1-7}
\multirow{3}{*}{CoT-1} & Easy & 0.755 $\pm$ 0.006 & 0.795 $\pm$ 0.007 & 0.661 $\pm$ 0.002 & 0.716 $\pm$ 0.009 & 0.059 $\pm$ 0.011 \\
 & Medium & 0.778 $\pm$ 0.009 & 0.718 $\pm$ 0.009 & 0.632 $\pm$ 0.013 & 0.627 $\pm$ 0.017 & 0.319 $\pm$ 0.013 \\
 & Hard & 0.765 $\pm$ 0.015 & 0.782 $\pm$ 0.016 & 0.660 $\pm$ 0.027 & 0.634 $\pm$ 0.021 & 0.363 $\pm$ 0.017 \\
\cmidrule(lr){1-7}
\multirow{2}{*}{CoT-2} & Medium & 0.844 $\pm$ 0.006 & 0.767 $\pm$ 0.009 & 0.670 $\pm$ 0.008 & 0.650 $\pm$ 0.014 & 0.361 $\pm$ 0.012 \\
 & Hard & 0.798 $\pm$ 0.011 & 0.841 $\pm$ 0.015 & 0.714 $\pm$ 0.012 & 0.685 $\pm$ 0.012 & 0.394 $\pm$ 0.034 \\
\cmidrule(lr){1-7}
\multirow{1}{*}{CoT-3} & Hard & 0.851 $\pm$ 0.014 & 0.881 $\pm$ 0.010 & 0.689 $\pm$ 0.008 & 0.720 $\pm$ 0.020 & 0.413 $\pm$ 0.015 \\
\bottomrule
\end{tabular}
\end{table}

\end{document}